\documentclass[letterpaper, 10 pt, conference]{ieeeconf}  

\IEEEoverridecommandlockouts                              

\usepackage{times} 
\usepackage{amsmath,amsfonts}
\usepackage{algorithmic}
\usepackage{algorithm}
\usepackage{array}
\usepackage{caption}
\usepackage{textcomp}
\usepackage{stfloats}
\usepackage{url}
\usepackage{verbatim}
\usepackage{graphicx}
\usepackage{cite}
\usepackage{hyperref}
\usepackage{tabularx}
\usepackage{multirow} 
\usepackage{subcaption} 
\usepackage{float}

\begin{document}
\title{\LARGE \bf Biomechanical Comparisons Reveal Divergence of Human and Humanoid Gaits}
\author{Luying Feng$^{1, 2}$, Yaochu Jin$^{2,*}$, \IEEEmembership{Fellow, IEEE}, Hanze Hu$^{3}$, Wei Chen$^{2}$ 
\thanks{$^{1}$Zhejiang University, Hangzhou, China.}
\thanks{$^{2}$School of Engineering, Westlake University, Hangzhou, China.}
\thanks{$^{3}$Ningbo Institute of Materials Technology and Engineering, Chinese Academy of Sciences, Ningbo, China.}
\thanks{$^{*}$Yaochu Jin is the corresponding author. E-mail: jinyaochu@westlake.edu.cn}
}

\maketitle
\thispagestyle{empty}
\pagestyle{empty}

\begin{abstract}
It remains challenging to achieve human-like locomotion in legged robots due to fundamental discrepancies between biological and mechanical structures. Although imitation learning has emerged as a promising approach for generating natural robotic movements, simply replicating joint angle trajectories fails to capture the underlying principles of human motion.
This study proposes a Gait Divergence Analysis Framework (GDAF), a unified biomechanical evaluation framework that systematically quantifies kinematic and kinetic discrepancies between humans and bipedal robots.
We apply GDAF to systematically compare human and humanoid locomotion across 28 walking speeds. To enable reproducible analysis, we collect and release a speed-continuous humanoid locomotion dataset from a state-of-the-art humanoid controller. We further provide an open-source implementation of GDAF, including analysis, visualization, and MuJoCo-based tools, enabling quantitative, interpretable, and reproducible biomechanical analysis of humanoid locomotion.
Results demonstrate that despite visually human-like motion generated by modern humanoid controllers, significant biomechanical divergence persists across speeds. Robots exhibit systematic deviations in gait symmetry, energy distribution, and joint coordination, indicating that substantial room remains for improving the biomechanical fidelity and energetic efficiency of humanoid locomotion.
This work provides a quantitative benchmark for evaluating  humanoid locomotion and offers data and versatile tools to support the development of more human-like and energetically efficient locomotion controllers.
The data and code will be made publicly available upon acceptance of the paper.
\end{abstract}

\section{INTRODUCTION}

The development of humanoid robots capable of natural, human-like locomotion has become a pivotal research objective with profound implications for human-robot coexistence. As humanoid platforms transition from controlled laboratory settings to real-world environments such as homes, workplaces, and public spaces, their gait quality directly impacts not only functional efficiency but also also public perception and acceptance. An unnatural mechanical gait can appear intimidating, generates annoying noise, and accelerates the wear of key foot components. Conversely, a fluid and biomimetic walking pattern not only extends operational endurance and enhances social acceptance but can also serve as an ideal platform for testing various rehabilitation and assistive devices, such as exoskeletons, in place of human subjects. However, achieving such naturalistic movement remains a fundamental challenge due to the complex interplay between mechanical design and control architecture. 

Reinforcement learning (RL) and imitation learning (IL) have enabled robots to achieve remarkable locomotion behaviors across diverse environments \cite{seo2025fasttd3}. For example, RL-based controllers have demonstrated robust sim-to-real walking on humanoids and adaptive locomotion across uneven terrains and planned footsteps \cite{singh2022learning}. However, despite their robustness and adaptability, these controllers often fail to produce natural humanlike motions. Motion imitation such as PHUMA \cite{lee2025phuma}, further enable learning humanlike movements. However, such policies often rely on simplistic joint-angle imitation without integrating biomechanical principles \cite{wehner2009optimizing}. Moreover, due to the lack of quantitative evaluation standards for robotic gait, reward tuning remains experience-driven and inefficient. Several studies have attempted to compare robotic and human locomotion, however, these evaluations are conducted at uncontrolled or isolated walking speeds. As walking speed fundamentally affects kinematic and kinetic properties, such isolated and inconsistent speed conditions significantly limit the completeness and interpretability of human–robot gait comparisons \cite{meng2018bipedal,kagami2004measurement,ji2021simulation}.

To address this gap, we propose a Gait Divergence Analysis Framework (GDAF) — a systematic, multi-dimensional framework designed to quantify and analyze the kinematic and kinetic discrepancies between human and humanoid locomotion. 
The contributions of our work are summarized as follows:

(1) We propose GDAF, a systematic biomechanical analysis framework for quantitatively comparing human and robotic gait patterns, including a comprehensive divergence index that aggregates multi-dimensional metrics.

(2) We systematically acquire and analyze gait patterns across 28 walking speeds and establish a publicly available benchmark dataset, analysis code and Visualization tools.

To the best of our knowledge, this is the first systematic, speed-varying biomechanical comparison between human locomotion and a humanoid robot trained with RL/IL. By introducing fine-grained speed increments of 0.05 m/s, our study not only captures the distinct biomechanical characteristics at 28 discrete walking speeds but also reveals subtle kinematic and kinetic variations across continuous speed transitions. This analysis provides benchmark for policy evaluation and offers detailed, multi-dimensional insights into the evolution of robot locomotion dynamics with respect to speed. The findings provide valuable guidance for robot gait training and advance policy development toward a new paradigm of evidence-based optimization.

\begin{figure*}[!t]
    \centering
    \includegraphics[width=0.95\linewidth]{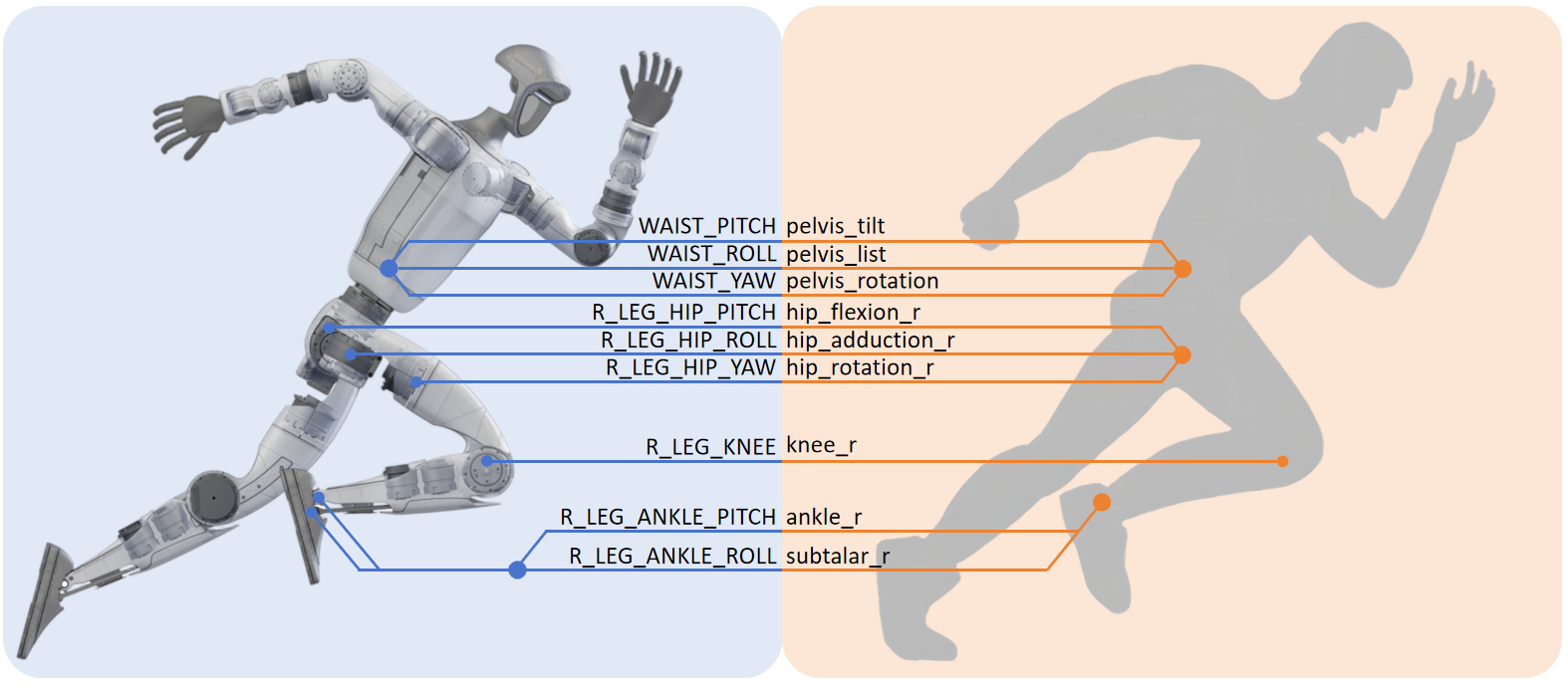}
    \caption{\label{joint_mapping}Joint mapping between humanoid robot and human.}
\end{figure*}

\section{Methodology}
\subsection{Data Acquisition and Processing}
\subsubsection{Human Gait Data Acquisition and Processing}
To facilitate a comparison between human and humanoid gait patterns, we utilized a large-scale open-source biomechanical gait database \cite{camargo2021comprehensive}. This database systematically captures biomechanical data from 22 able-bodied adults (age: 21 ± 3.4 yr, height: 1.70 ± 0.07 m, mass: 68.3 ± 10.83 kg) across various locomotion modes, including level-ground walking, treadmill walking, stair ascent/descent, and ramp ascent/descent. We focused on treadmill walking data, which encompasses 28 walking speeds ranging from 0.5 m/s to 1.85 m/s in 0.05 m/s increments, with each speed maintained for 30 seconds.

We processed the raw biomechanical data using custom open-source code integrated with the OpenSim platform \cite{delp2007opensim}, implementing a standardized pipeline where motion capture and force plate inputs underwent inverse kinematics calculations to derive joint angles, followed by inverse dynamics computations to determine joint moments. Joint power was subsequently calculated through numerical differentiation of joint angles multiplied by corresponding joint moments, with all kinetic outputs (moments and powers) normalized to subject body mass (Nm/kg and W/kg). To enable cross-stride comparisons, gait cycles were defined from 0\% to 100\% based on heel-strike events. All biomechanical data points were then temporally aligned through linear interpolation according to gait cycle percentage. Heel-strike events were identified as the points where the linear velocity of the right heel marker in the motion capture data reached zero.

\subsubsection{Robotic Gait Data Acquisition and Processing}
The Unitree G1 humanoid robot was selected for this study for its superior hardware capabilities. For systematic gait comparison between human and humanoid locomotion, we utilized Unitree's proprietary walking/running motion controller - trained combined with RL and IL and recognized as one of the most advanced and stable locomotion models available. 

The experiments recorded the following data from all joints across 28 speed conditions ranging from 0.5 m/s to 1.85 m/s in 0.05 m/s increments: joint angles, angular velocities, and estimated torques. The data collection frequency is 200 Hz, which matches the acquisition frequency used in the above databases for human movement. Each walking/running velocity was held for at least 25 s to capture steady-state walking. For gait cycle analysis, acceleration and deceleration phases were excluded to focus solely on steady-state walking. To facilitate comparative analysis and mitigate the effects of gait variability, we adopted the same methodology used in human gait studies: gait cycles were normalized from 0\% to 100\% based on right heel-strike events, with averaging performed across multiple cycles at the same speed. Due to the lack of motion capture and plantar pressure data, the timing of abrupt changes in ankle pitch angle was instead utilized as a surrogate indicator for gait phase division. Joint powers were calculated by angular velocities and torques, and the joint moments and powers will then normalized to robot body mass to enable cross-stride comparisons.

It is important to note a methodological distinction between human and robot data acquisition: human data were collected on a treadmill, while robot data were collected over level ground. Although previous studies have documented biomechanical differences between treadmill and overground walking in humans \cite{lee2008biomechanics}, this experimental design was necessitated by practical constraints. Humans cannot maintain precise, arbitrary walking speeds (e.g., exactly 1.0 m/s) during level-ground walking; existing open-source datasets typically record overground gait at only three self-selected speeds (slow, normal, fast) relative to each individual's preferred pace. Conversely, treadmill walking with the Unitree G1 was deemed unsafe due to lateral drift in the absence of visual feedback.

\subsubsection{Joint Mapping}
Fig. \ref{joint_mapping} establishes the biomechanical correspondence between human joints and robotic actuators. The mapping was determined based on anatomical function and movement planes. Note that the metatarsophalangeal joint has no direct robotic equivalent and was excluded from this comparison. 
Additionally, since humans and robots employ different coordinate systems, we transformed the human motion data into the robot's coordinate frame prior to analysis to ensure consistent comparison.

\subsection{A Gait Divergence Analysis Framework}
The core of this study is to propose and apply GDAF, which systematically quantifies similarities and discrepancies between human and humanoid gait patterns across different walking speeds. Built on the unified preprocessing pipeline described above, GDAF operates on normalized gait cycles and harmonized joint coordinates, and consists of three stages: (i) Data Preparation and Visualization, (ii) Multi-dimensional Quantitative Divergence Analysis, and (iii) Comprehensive GDAF Cost. 
The metrics used in GDAF are inspired by and adapted from established gait biomechanics analysis methods. Waveform similarity is quantified using Pearson correlation, a standard measure for comparing joint kinematic and kinetic trajectories \cite{winter2009biomechanics}. Bilateral symmetry is also a widely used indicator of interlimb coordination and gait quality \cite{patterson2012gait, alves2020quantifying}. Energetic behavior reflects how mechanical energy is generated, absorbed, and transferred across joints during locomotion, providing insight into propulsion strategies and joint functional roles \cite{winter1983energy}. Torque–angle loops further characterize joint mechanical behavior and energy exchange properties \cite{rouse2014clutchable}.

\subsubsection{Data Preparation and Visualization}

This section establishes a standardized data preparation schema to enhance the generality and reusability of the analysis code. By ensuring that all subsequent code modules operate on this unified format, the computational framework of this study can be easily adapted and migrated to other data source, such as datasets collected from different robots under different speed conditions, or data acquired from simulation environments.

Concretely, all processed biomechanical data are exported into a unified \texttt{.mat} schema. Each dataset is stored as a set of channel–speed matrices and meta-data:
\texttt{pos[channel, speed]} contains joint-angle trajectories in degrees; \texttt{torque[channel, speed]} and \texttt{power[channel, speed]} store body-mass-normalized joint moments (Nm/kg) and powers (W/kg), respectively; \texttt{channel\_label} lists the anatomical joint names; and \texttt{unique\_speeds} records the walking speeds in m/s. Within each cell, time series are resampled to a normalized gait cycle from 0\% to 100\% and averaged over multiple strides, so that any downstream analysis can operate purely in the gait-percentage domain without needing to access raw motion-capture or motor data. Building on this standardized format, we provide open-source notebooks that reproduce the figures and analyses presented in this paper: \texttt{GDAF\_01\_Data\_Visualization.ipynb} for data visualization and \texttt{GDAF\_02\_Divergence\_Analysis.ipynb} for divergence analysis. 

To complement these quantitative tools, we also release a custom MuJoCo-based visualization tool (\texttt{mujoco\_gait\_player.py}) for convenient data inspection and flexible gait comparison.
A screenshot of the visualization interface is shown in Fig. \ref{fig:mujoco_gait_viewer}.
The tool loads the preprocessed human and robot gait datasets (joint position, torque, and power profiles across multiple walking speeds) and drives a 29-dof Unitree G1 model in a MuJoCo simulation.
We fix the floating-base pose and replay the joint trajectories at selected speeds, which allows side-by-side inspection of human and robotic motions as well as rapid debugging of joint mappings and kinematic constraints.

Users can apply GDAF to their own data by exporting joint angles, moments, and powers into the same \texttt{.mat} schema and updating file paths in the provided code.

\subsubsection{Multi-dimensional Quantitative Divergence Analysis}

Building on the unified data representation, GDAF quantitatively compares human and robotic gait across three complementary dimensions: waveform similarity, bilateral symmetry and energetic behavior. All analyses are performed per joint (or per bilateral joint pair where applicable) and per walking speed. 

First, we assess the waveform similarity. For each joint $j$ and speed $v$, we extract the human and robot trajectories of joint angle, joint moment, and joint power over the gait cycle,
\[
\mathbf{x}^{\text{H}}_{j,v}(k),\ \mathbf{x}^{\text{R}}_{j,v}(k), \quad k = 1,\dots, N,
\]
where $N$ is the number of time-normalized samples. We quantify waveform similarity using the Pearson correlation coefficient,
\[
r_{j,v} = \frac{ \sum_{k=1}^N \left( x^{\text{H}}_{j,v}(k) - \bar{x}^{\text{H}}_{j,v} \right) \left( x^{\text{R}}_{j,v}(k) - \bar{x}^{\text{R}}_{j,v} \right) }{ \sqrt{ \sum_{k=1}^N \left( x^{\text{H}}_{j,v}(k) - \bar{x}^{\text{H}}_{j,v} \right)^2 } \sqrt{ \sum_{k=1}^N \left( x^{\text{R}}_{j,v}(k) - \bar{x}^{\text{R}}_{j,v} \right)^2 } },
\]

This procedure is applied separately to joint angles, moments, and powers, yielding a set of speed-dependent similarity metrics for each biomechanical quantity: $r^{\text{moment}}_{j,v}$, and $r^{\text{power}}_{j,v}$.  The combined waveform similarity at speed $v$ is
\[
R^{\text{wav}}_v = 0.5\,\bar{r}^{\text{angle}}_v + 0.3\,\bar{r}^{\text{moment}}_v + 0.2\,\bar{r}^{\text{power}}_v,
\]

where $\mathcal{J}$ denotes the set of mapped joints, $\bar{r}^{\text{angle}}_v = \frac{1}{|\mathcal{J}|}\sum_{j \in \mathcal{J}} r^{\text{angle}}_{j,v}$ is the mean angle similarity over all joints, and $\bar{r}^{\text{moment}}_v$, $\bar{r}^{\text{power}}_v$ are defined analogously. The weights (0.5, 0.3, 0.2) reflect that joint angles are typically the most stable and repeatable in gait , moments are derived from inverse dynamics and tend to be noisier, and power (product of moment and angular velocity) is more sensitive to phase and magnitude errors; this prioritization is consistent with the common practice in clinical gait analysis of emphasizing kinematic consistency before kinetics. A larger $R^{\text{wav}}_v$ indicates greater overall waveform agreement between human and robot at that speed.

Second, we quantify bilateral symmetry using feature point comparison. For each bilateral joint pair $p$ (e.g., left and right hip flexion) and speed $v$, let $\theta^{\text{L}}_{p,v}(k)$ and $\theta^{\text{R}}_{p,v}(k)$ denote the left and right-side trajectories over the normalized gait cycle $k = 1,\dots,N$. We extract scalar features: maximum flexion $\phi_{\max}$ and maximum extension $\phi_{\min}$:
\[
\phi^{\text{L}}_{\max} = \max_k \theta^{\text{L}}_{p,v}(k),\qquad \phi^{\text{L}}_{\min} = \min_k \theta^{\text{L}}_{p,v}(k),
\]
and analogously for the right side. Denote by $\phi^{\text{L},E}_{\max}$, $\phi^{\text{R},E}_{\max}$, $\phi^{\text{L},E}_{\min}$, $\phi^{\text{R},E}_{\min}$ the features for entity $E \in \{\text{H}, \text{R}\}$ (human or robot). The bilateral symmetry index (SI) combines the absolute differences of these two features (maximum and minimum), with a  human-reference denominator for cross-entity comparability:
\[
\text{SI}^{E}_{p,v} = \frac{2\left( \bigl| \phi^{\text{L},E}_{\max} - \phi^{\text{R},E}_{\max} \bigr| + \bigl| \phi^{\text{L},E}_{\min} - \phi^{\text{R},E}_{\min} \bigr| \right)}{ \bigl( \bigl| \phi^{\text{L},\text{H}}_{\max} \bigr| + \bigl| \phi^{\text{R},\text{H}}_{\max} \bigr| \bigr) + \bigl( \bigl| \phi^{\text{L},\text{H}}_{\min} \bigr| + \bigl| \phi^{\text{R},\text{H}}_{\min} \bigr| \bigr) + \epsilon },
\]
where the denominator uses human feature magnitudes as the reference scale. This ensures human and robot SI values are directly comparable (higher SI indicates greater asymmetry of that entity relative to human motion scale). The per-speed combined symmetry index over the six lower-limb bilateral pairs can be expressed by
\[
\bar{\text{SI}}^{E}_v = \frac{1}{|\mathcal{P}|}\sum_{p \in \mathcal{P}} \text{SI}^{E}_{p,v},
\]
where $\mathcal{P}$ denotes the set of six bilateral pairs. A larger $\bar{\text{SI}}^{E}_v$ indicates greater overall bilateral asymmetry of that entity at speed $v$.

Third, we characterize energetic behavior. To describe differences in mechanical work, we decompose the joint power profiles into positive and negative components and numerically integrate over the gait cycle,
\[
\begin{gathered}
W^{+}_{j,v} = \int_0^{T} \max\bigl(P^{\text{H}}_{j,v}(t), 0\bigr)\, dt, \\
W^{-}_{j,v} = \int_0^{T} \min\bigl(P^{\text{H}}_{j,v}(t), 0\bigr)\, dt,
\end{gathered}
\]
and analogously obtain $W^{+,\text{R}}_{j,v}$ and $W^{-,\text{R}}_{j,v}$ for the robot. From these we derive two aggregate metrics. \textbf{(a) Bilateral work symmetry} over the six lower-limb joint pairs: for each bilateral pair $p \in \mathcal{P}$ and entity $E \in \{\text{H}, \text{R}\}$, let $W^{+,\text{L},E}_{p,v}$, $W^{-,\text{L},E}_{p,v}$ and $W^{+,\text{R},E}_{p,v}$, $W^{-,\text{R},E}_{p,v}$ denote the positive and negative work on the left and right side. With a human-reference denominator for cross-entity comparability.
\[
\begin{split}
A^{\text{W},E}_{p,v} &= \frac{1}{2}\biggl( \frac{2\bigl|W^{+,\text{L},E}_{p,v} - W^{+,\text{R},E}_{p,v}\bigr|}{\bigl|W^{+,\text{L},\text{H}}_{p,v}\bigr| + \bigl|W^{+,\text{R},\text{H}}_{p,v}\bigr| + \epsilon}\\
&\quad + \frac{2\bigl|W^{-,\text{L},E}_{p,v} - W^{-,\text{R},E}_{p,v}\bigr|}{\bigl|W^{-,\text{L},\text{H}}_{p,v}\bigr| + \bigl|W^{-,\text{R},\text{H}}_{p,v}\bigr| + \epsilon} \biggr).
\end{split}
\]

The per-speed combined work symmetry index is $\bar{A}^{\text{W},E}_v = \frac{1}{|\mathcal{P}|}\sum_{p \in \mathcal{P}} A^{\text{W},E}_{p,v}$. A higher $\bar{A}^{\text{W},E}_v$ indicates greater left-right work asymmetry of that entity at speed $v$. \textbf{(b) Human-robot work divergence} across all joints: for each joint $j$ and speed $v$, the per-joint work divergence is
\[
d^{\text{work}}_{j,v} = \frac{1}{2}\left( \frac{\bigl|W^{+,\text{H}}_{j,v} - W^{+,\text{R}}_{j,v}\bigr|}{\bigl|W^{+,\text{H}}_{j,v}\bigr| + \epsilon} + \frac{\bigl|W^{-,\text{H}}_{j,v} - W^{-,\text{R}}_{j,v}\bigr|}{\bigl|W^{-,\text{H}}_{j,v}\bigr| + \epsilon} \right).
\]
The per-speed human--robot work divergence over all mapped joints is $\bar{d}^{\text{work}}_v = \frac{1}{|\mathcal{J}|}\sum_{j \in \mathcal{J}} d^{\text{work}}_{j,v}$. A higher $\bar{d}^{\text{work}}_v$ indicates greater overall work difference between human and robot at that speed.

\subsubsection{Comprehensive GDAF Cost}
The comprehensive GDAF cost combines robot symmetry $\mathcal{S}^{\text{R}}_v$ and human-likeness  $\mathcal{H}_v$.
\textbf{a) Gait symmetry.} We combine bilateral symmetry $\bar{\text{SI}}^{\text{R}}_v$ and work symmetry $\bar{A}^{\text{W},\text{R}}_v$, scaling the work-asymmetry term by $1/10$ so that kinematic symmetry dominates:
\[
\mathcal{S}^{\text{R}}_v = 0.5\,\bar{\text{SI}}^{\text{R}}_v + 0.5\cdot\frac{\bar{A}^{\text{W},\text{R}}_v}{10}.
\]

This balances kinematic left–right consistency and energetic left–right symmetry, mitigating the interference caused by their disparate units.
\textbf{b) Human-likeness.} We combine waveform similarity $R^{\text{wav}}_v$ and work divergence $\bar{d}^{\text{work}}_v$ as $\mathcal{H}_v$. Since $R^{\text{wav}}_v$ is a similarity (higher = better), we use $1 - R^{\text{wav}}_v$ as the cost. The work-divergence term is scaled by $1/10$ so as to mitigate the interference caused by their disparate units:
\[
\mathcal{H}_v = \frac{1}{2}\left( (1 - R^{\text{wav}}_v) + \frac{\bar{d}^{\text{work}}_v}{10} \right).
\]

\textbf{3) Comprehensive index.} We combine robot symmetry and human-likeness with equal weight:
\[
C^{\text{GDAF}}_v = 0.5\,\mathcal{S}^{\text{R}}_v + 0.5\,\mathcal{H}_v.
\]

Lower $C^{\text{GDAF}}_v$ indicates better robot symmetry and human-likeness at speed $v$. 

\section{Results}
\subsection{Data Visualization}
\begin{figure}[tbp]
    \centering
    \includegraphics[width=0.9\linewidth]{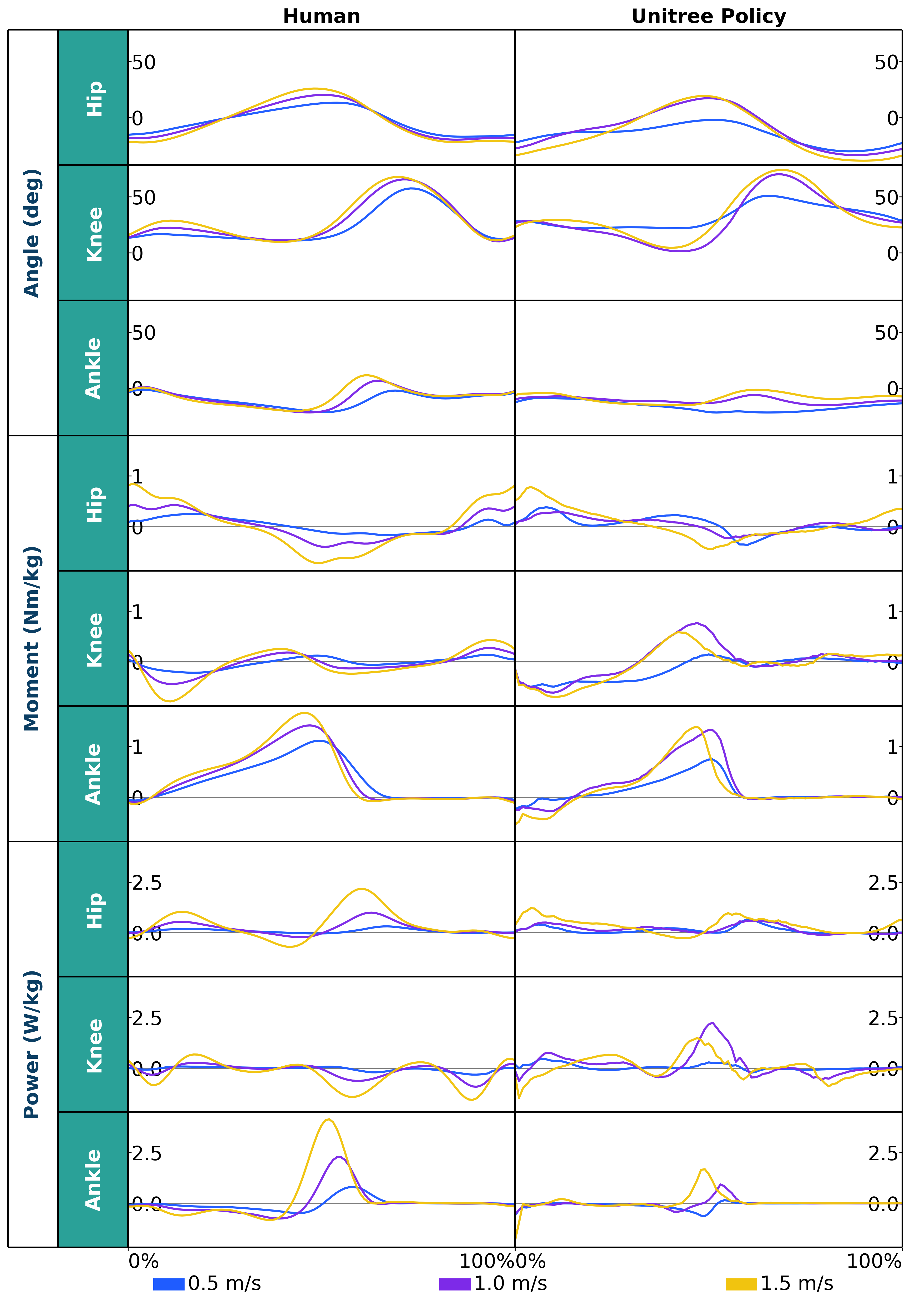}
    \caption{Comparison of lower-limb joint angles, moments, and power between human subjects and humanoid robot at three walking speeds.}
    \label{duibi}
\end{figure}
As shown in Fig. \ref{fig:human_robot_joint_comparison}, we visualize lower-limb joint angles, moments and powers for human subjects and the humanoid across 28 steady-state walking speeds. The figure reveals a number of interesting phenomena. Overall, as speed increases, human joint profiles exhibit smoothly increasing peak magnitudes with relatively compliant, rounded waveforms. In contrast, although the robot's joint curves follow similar variation patterns with increasing speed, their waveforms appear less smooth and rounded, with less uniform amplitude modulation, resulting in a somewhat irregular appearance.

We can examining individual joints to yield further insights. For the left hip joint, both the human and robot exhibit remarkably similar angle trajectories over the gait cycle, with consistent speed-dependent modulation patterns. This similarity also extends to joint moment profiles. Although the joint power profiles differ appreciably in shape, at 1.85 m/s both operate within a comparable range of approximately -1 to 3 W/kg; however, at lower speeds, the robot consistently performs more positive work at this joint. The right ankle angle reveals a different pattern. Human ankle dorsiflexion–plantarflexion trajectories maintain highly consistent waveform shapes across all 28 speeds, with smoothly increasing peak magnitudes and slightly advancing peak timing as speed increases. In contrast, the robot's ankle angle profiles approximate human-like morphology only at higher speeds; at lower speeds, the characteristic peak observed in human gait diminishes substantially, or even disappears entirely.

To enhance clarity, we include Fig. \ref{duibi}, which juxtaposes the gait profiles of human subjects and the Unitree robot across three walking speeds (0.5 m/s, 1.0 m/s, and 1.5 m/s) over a normalized gait cycle (0–100\%). While the knee joint angle trajectory of the robot follows a similar pattern to that of humans, its knee torque and power are substantially higher. 

\begin{figure*}[htbp]
    \centering
    \includegraphics[width=0.75\linewidth]{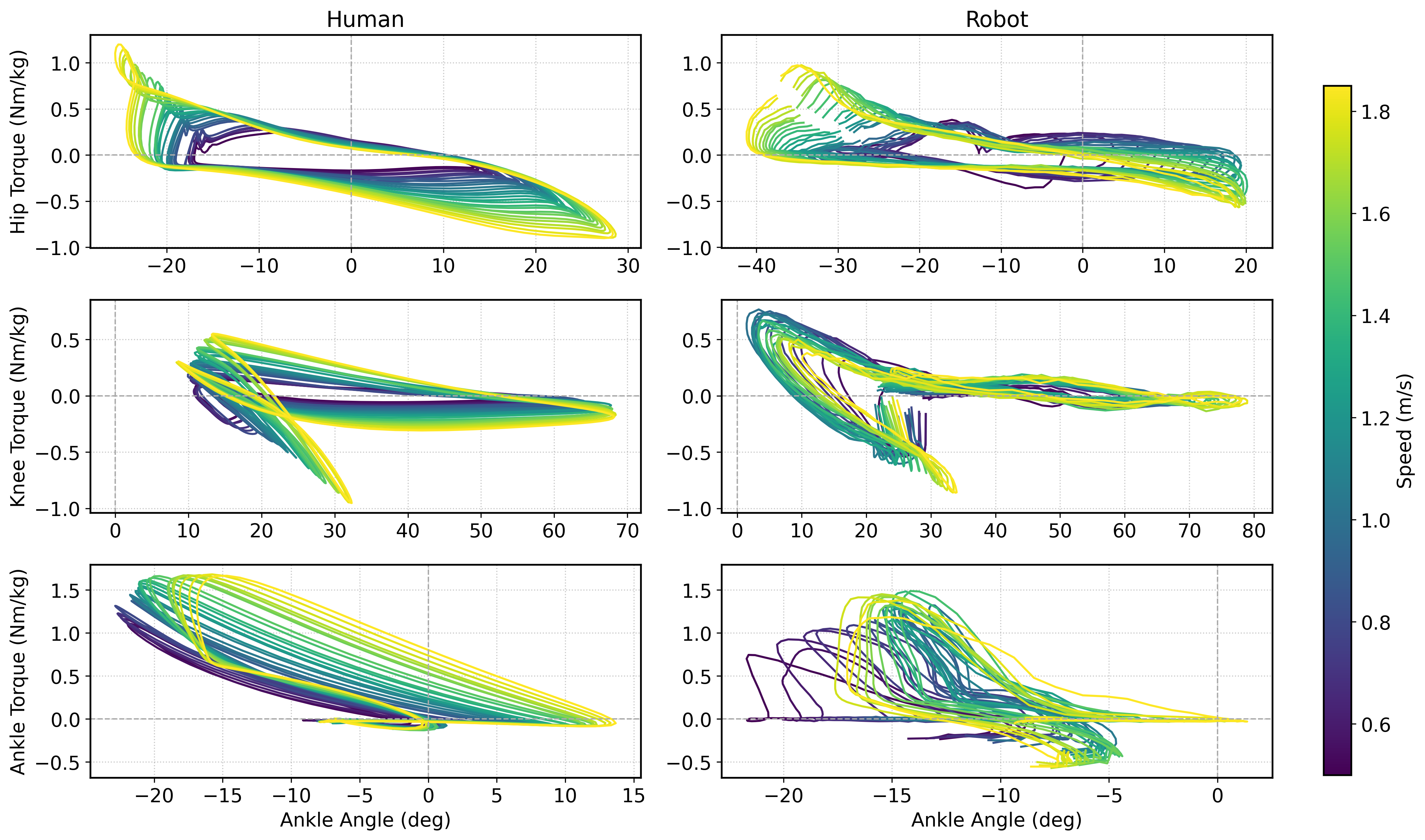}
    \caption{Comparison of torque-angle loops between human subjects and humanoid robot at three walking speeds.}
    \label{torque_angle_loops}
\end{figure*}
Figure \ref{torque_angle_loops} illustrates the joint torque–angle loops of the hip, knee and ankle across 28 walking speeds (0.5–1.8 m/s) for human and humanoid. The color gradient denotes speed, and each closed curve corresponds to gait cycle under certain speed. Human loops are smooth and show a clear energy storage–release pattern, the robot's loops exhibit low similarity to human patterns, appearing distorted and varying irregularly with speed. These qualitative observations motivate the subsequent GDAF-based quantitative divergence analysis, which systematically assesses where and how the robot’s kinematic and kinetic patterns deviate from human biomechanics across joints and speeds.

\subsection{Gait Divergence Analysis}
\subsubsection{Waveform Similarity}
\begin{figure}[!t]
    \centering
    \begin{subfigure}{\linewidth}
        \centering
        \includegraphics[width=0.92\linewidth]{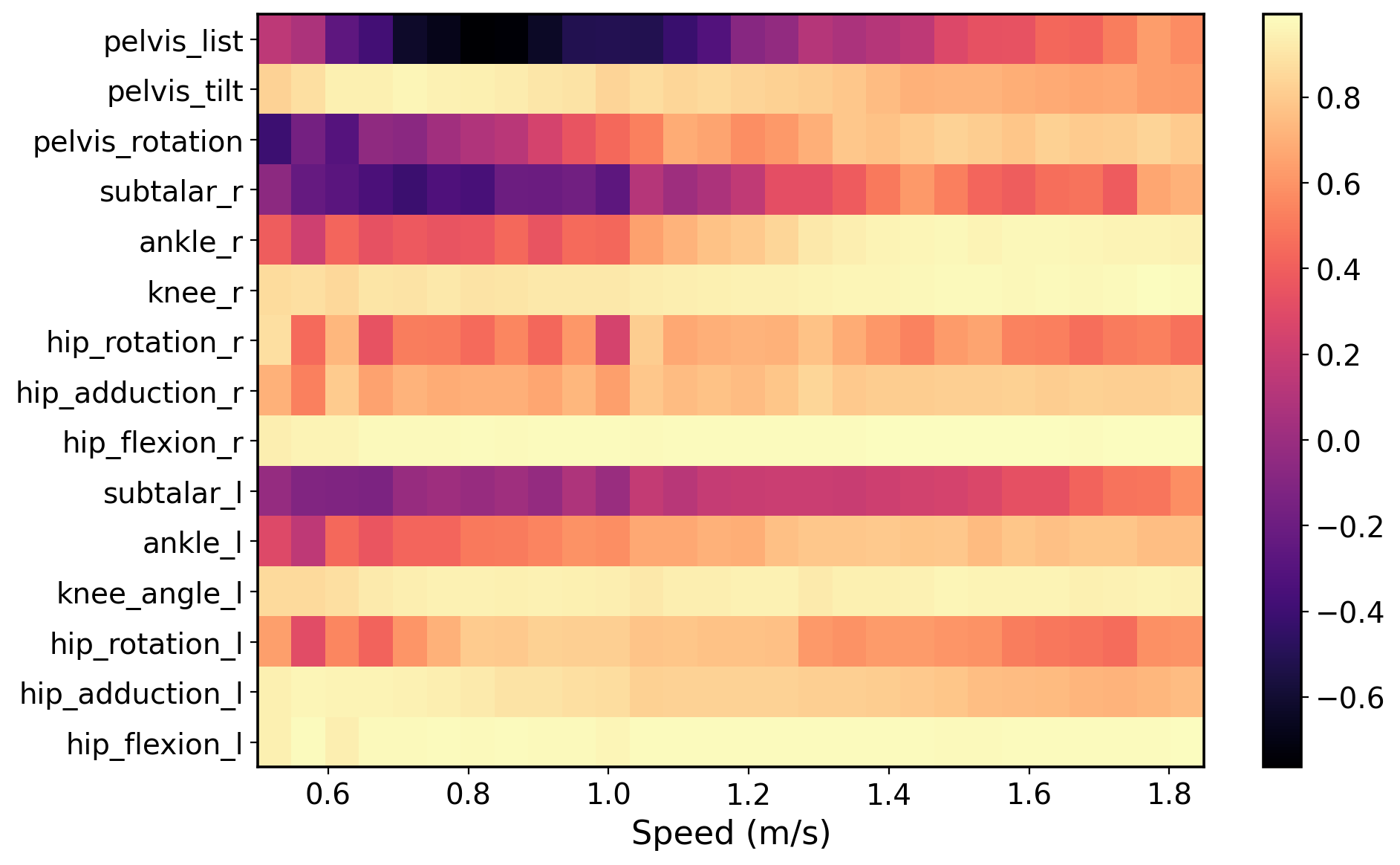}
        \caption{Joint angle similarity across speeds.}
        \label{fig:angle_similarity_heatmap}
    \end{subfigure}
    \begin{subfigure}{\linewidth}
        \centering
        \includegraphics[width=0.92\linewidth]{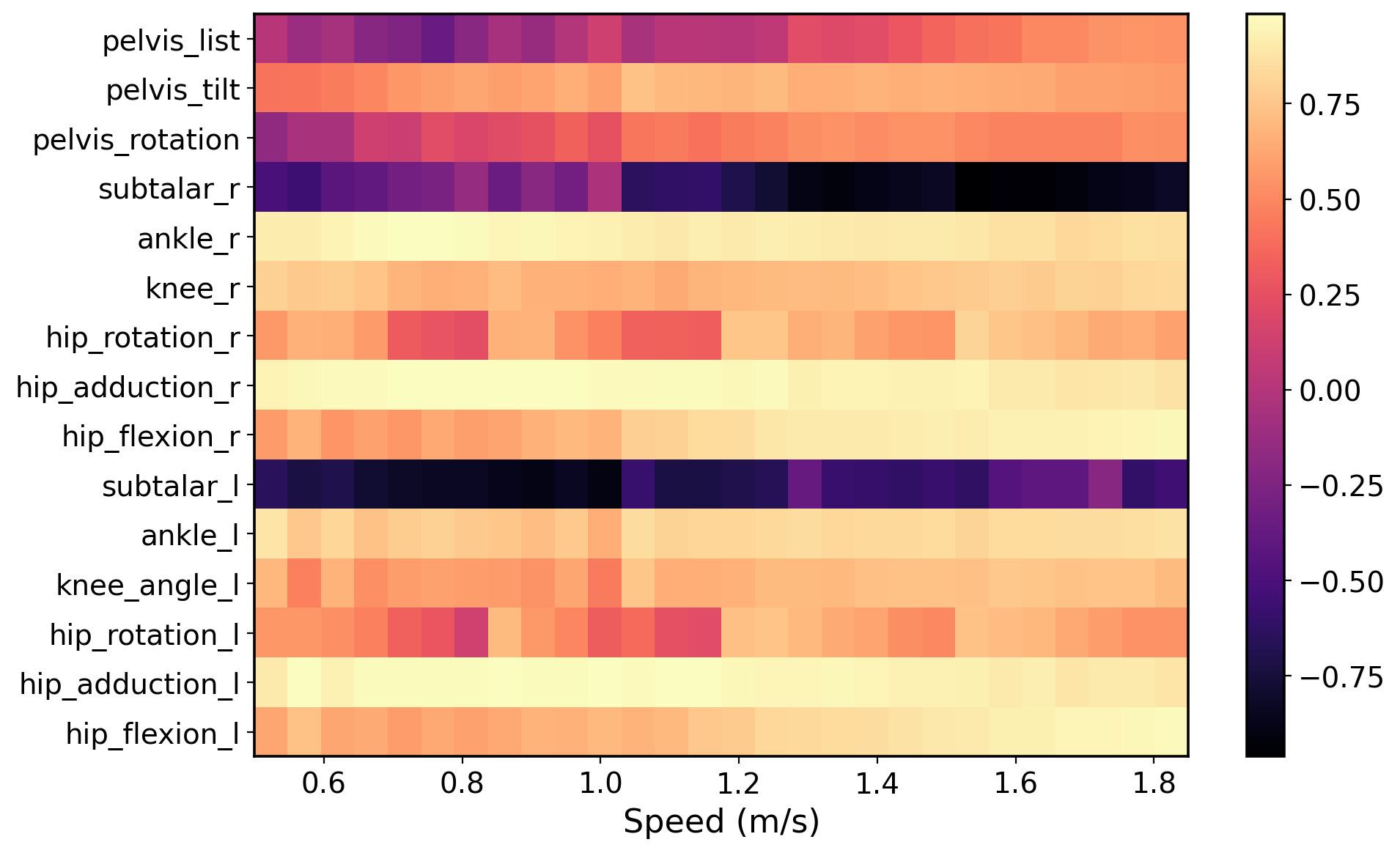}
        \caption{Joint moment similarity across speeds.}
        \label{fig:moment_similarity_heatmap}
    \end{subfigure}
    \begin{subfigure}{\linewidth}
        \centering
        \includegraphics[width=0.92\linewidth]{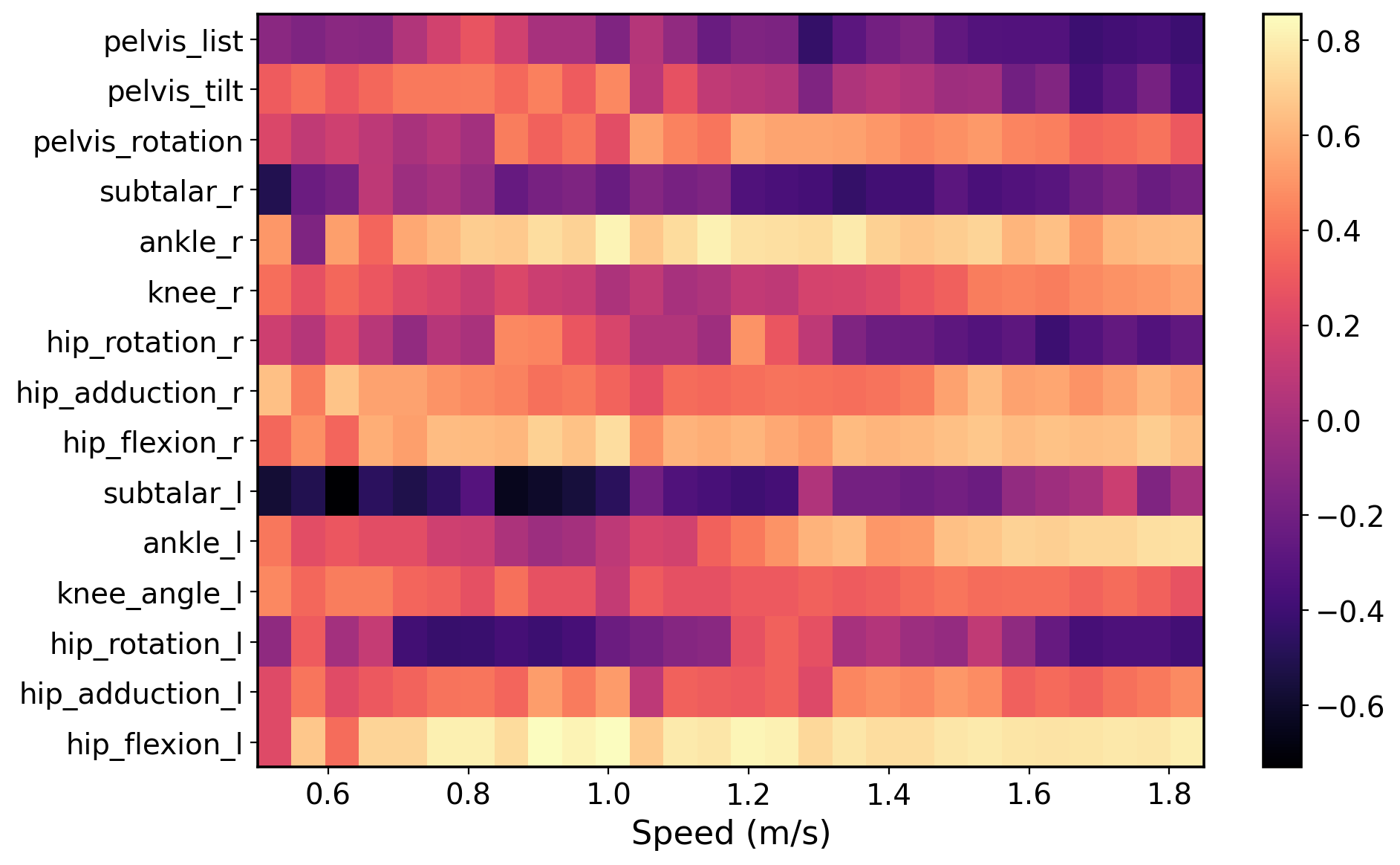}
        \caption{Joint power similarity across speeds.}
        \label{fig:power_similarity_heatmap}
    \end{subfigure}
    \caption{Multidimensional similarity heatmaps of joint angles, moments, and power across walking speeds}
    \label{fig:similarity_heatmaps}
\end{figure}

The similarity heatmaps in Fig. \ref{fig:similarity_heatmaps} visualize the distribution of correlation coefficients between robotic and human gait profiles across 28 walking speeds. The three subfigures correspond to the similarity of joint angle, moment and power, respectively. The heatmaps reveal that hip and knee joint angles achieve consistently high similarity (correlation coefficients typically above 0.8) between human and robot across most walking speeds. In contrast, joint moment and power similarities exhibit more pronounced variations, with both side of subtalar joint showing noticeably lower similarity. Additionally, for several joints (pelvis list, pelvis rotation, subtalar and ankle), similarity tends to increase progressively with walking speed.

\subsubsection{Bilateral Symmetry}
\begin{figure*}[tbp]
    \centering
    \includegraphics[width=0.77\linewidth]{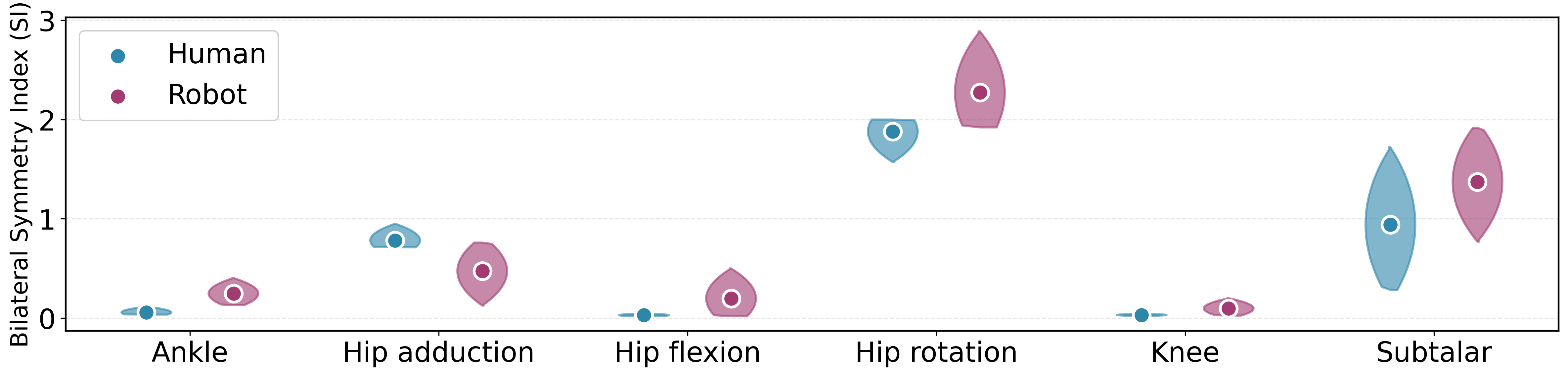}
    \caption{Bilateral symmetry index (SI) across joint pairs for human and robot.}
    \label{bilateral_symmetry}
\end{figure*}
Figure. \ref{bilateral_symmetry} presents the distribution of symmetry index (SI) values across all walking speeds for six major joint pairs. 
In the human data, near-symmetric behavior is evident in sagittal-plane dominant joints, particularly hip flexion and the knee, where SI values remain close to zero with minimal dispersion across speeds. This reflects highly coordinated left–right kinematics and stable gait periodicity. By contrast, the robot exhibits systematically higher SI values for these joints, indicating greater inter-limb discrepancies in both range of motion and trajectory extrema.
The divergence becomes even more pronounced in distal joints. At the subtalar and ankle, the robot shows substantially higher SI compared to humans, highlighting amplified asymmetry in regions critical for foot placement and ground contact.

\subsubsection{Energetic Divergence}
\begin{figure*}[htbp]
    \centering
    \includegraphics[width=0.77\linewidth]{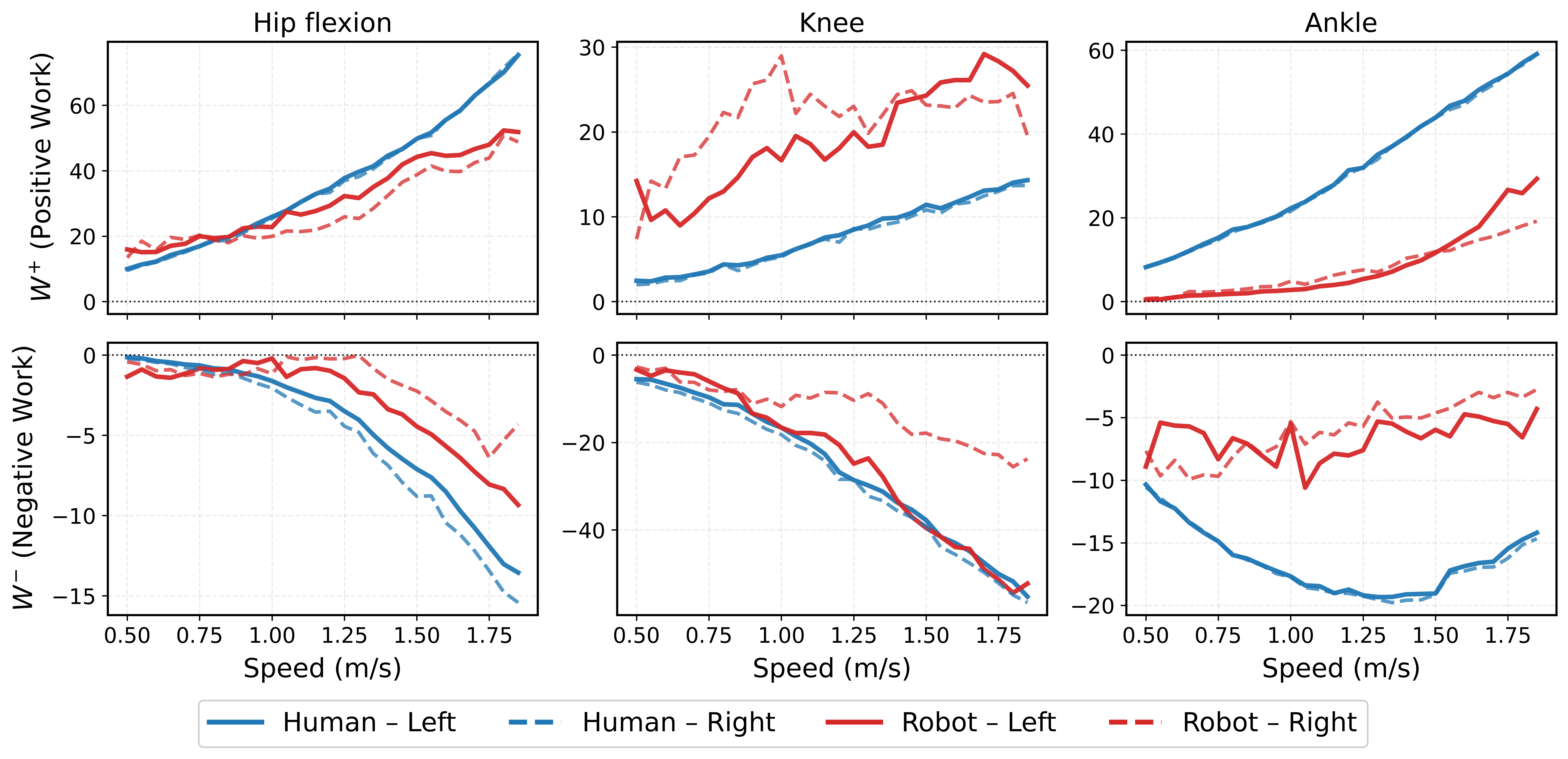}
    \caption{Positive/negative work from power profiles.}
    \label{energy}
\end{figure*}

The GDAF framework also quantifies energetic divergence between human and humanoid gaits. Figure \ref{energy} summarizes the positive and negative joint work ($W^+$ and $W^-$) for the hip, knee, and ankle, obtained by integrating the joint power profiles across walking speeds from 0.5 to 1.8~m/s. For human gait, the positive work at the hip and ankle increases monotonically with speed, whereas the knee produces relatively small positive work with only a mild speed dependence. In terms of negative work, the human hip and knee $W^-$ decrease continuously as speed increases (i.e., progressively stronger energy absorption), while the ankle $W^-$ shows a non-monotonic trend, decreasing first and then increasing with speed.

Compared with humans, the robot exhibits a clear redistribution of mechanical work. The robot generates substantially larger knee positive work across the entire speed range, indicating a knee-dominant power-generation strategy. In contrast, the robot ankle positive work is markedly lower than that of humans, suggesting insufficient ankle push-off. At the hip, robot positive work also increases with speed but remains lower than the human hip at higher speeds, implying that the robot does not fully compensate for the reduced ankle $W^+$ by increasing hip output.
On the absorption side, the magnitude of robot hip negative work is generally smaller than that of humans, and robot ankle negative work remains small and weakly sensitive to speed. Regarding inter-limb consistency, human left/right curves largely overlap across joints and speeds, indicating high bilateral consistency. The robot shows more pronounced left--right differences across multiple joints, particularly in knee negative work, suggesting a more asymmetric bilateral distribution of joint work, most evident at low-to-moderate speeds.

\subsubsection{Comprehensive Evaluation}

\begin{table*}[t]
    \centering
    \caption{Comprehensive GDAF indices per speed (m/s). Symmetry rows: robot value (human reference in parentheses).}
    \label{tab:gdaf}
    \begin{tabular}{llccccc}
    \hline
    \textbf{Category} & \textbf{Metric} & 0.5 & 0.75 & 1.0 & 1.25 & 1.5 \\
    \hline
      \multirow{3}{*}{Symmetry} & $\mathrm{SI}^{\mathrm{R}}_v$ ($\mathrm{SI}^{\mathrm{H}}_v$) & 0.8292 (0.8067) & 0.8488 (0.7259) & 0.8254 (0.6252) & 0.7511 (0.5545) & 0.6772 (0.4936) \\
       & $A^{\mathrm{W,R}}_v$ ($A^{\mathrm{W,H}}_v$) & 3.3708 (0.3127) & 0.9428 (0.2172) & 0.9911 (0.1416) & 0.6542 (0.1870) & 0.2507 (0.1355) \\
       & $\mathcal{S}^{\mathrm{R}}_v$ &\textbf{ 0.5831} &\textbf{ 0.4715} & \textbf{0.4623} & \textbf{0.4083 }& \textbf{0.3511} \\
      \multirow{3}{*}{Human-likeness} & $R^{\mathrm{wav}}_v$ & 0.4316 & 0.4182 & 0.4458 & 0.5668 & 0.5856 \\
       & $d^{\mathrm{work}}_v$ & 3.0759 & 1.1461 & 0.9081 & 0.9929 & 0.5966 \\
       & $\mathcal{H}_v$ & \textbf{0.4380} & \textbf{0.3482} & \textbf{0.3225} & \textbf{0.2662} & \textbf{0.2371} \\
      \multirow{1}{*}{Comprehensive} & $C^{\mathrm{GDAF}}_v$ & \textbf{0.5106} &\textbf{ 0.4099} & \textbf{0.3924 }& \textbf{0.3373} & \textbf{0.2941} \\
    \hline
    \end{tabular}
\end{table*}

\autoref{tab:gdaf} reports the comprehensive indices at representative speeds. Overall, the comprehensive score $C^{\mathrm{GDAF}}_v$ decreases substantially with speed, from 0.5106 at 0.5 m/s to 0.2941 at 1.5 m/s, indicating progressively improved gait quality as speed increases. This trend is jointly driven by reductions in both pillars: the robot-symmetry term $\mathcal{S}^{\mathrm{R}}_v$ drops from 0.5831 to 0.3511, and the human-likeness term $\mathcal{H}_v$ drops from 0.4380 to 0.2371. Notably, across all reported speeds, humans consistently show better bilateral symmetry than the robot in both measures: $\mathrm{SI}$ and $A$, with the largest disparity appearing in work symmetry at low speed.

\section{Discussion}

This work proposed GDAF as a unified biomechanical evaluation framework for quantifying gait divergence between humans and humanoid under matched walking speeds. By integrating waveform similarity, bilateral symmetry, and energetic behavior into a single pipeline and a comprehensive cost, GDAF enables both joint-level diagnosis and speed-dependent trend analysis. Applying GDAF to human data and self collected Unitree G1 data across 28 speeds reveals several consistent and practically actionable observations.

\subsection{Speed-dependent gait quality and the slow-speed bottleneck}
A key result is the monotonic improvement of the robot gait with increasing speed, reflected by the decreasing comprehensive score $C^{\mathrm{GDAF}}_v$ (Table \ref{tab:gdaf}). This improvement is jointly driven by reduced asymmetry ($\mathcal{S}^{\mathrm{R}}_v$) and increased human-likeness ($\mathcal{H}_v$). Notably, the largest disparity appears at low speed, especially in the work symmetry term $A^{\mathrm{W,R}}_v$, suggesting that slow walking remains a bottleneck regime where the controller struggles to maintain consistent bilateral coordination and smooth energetic modulation. From a control perspective, slow walking often requires precise balance regulation with smaller momentum, which imply that evaluating humanoid locomotion only at medium or fast speeds may systematically overestimate its overall biomechanical fidelity.

\subsection{Proximal-to-distal gradient in human--robot agreement}
The similarity heatmaps indicate a clear proximal-to-distal gradient: hip and knee angles achieve high agreement across most speeds, whereas distal joints (ankle and subtalar) exhibit lower similarity, particularly in kinetics (moments and powers). This gradient is consistent with the qualitative torque--angle loops (Fig.~\ref{torque_angle_loops}), where human loops show smooth, self-similar scaling with speed while robot loops deform irregularly. Distal joints are strongly shaped by contact dynamics and by biological elastic elements; in humans, ankle push-off leverages tendon elasticity and passive energy storage--release that naturally produces smooth quasi-stiffness patterns. In contrast, a rigid-actuated humanoid must actively synthesize similar behaviors through control and foot--ground interaction, which is more sensitive to timing and contact constraints. Therefore, distal-joint divergence should be interpreted not only as a control limitation but also as a combined outcome of actuation, foot design, and contact modeling.

\subsection{Redistribution of mechanical work and a knee-dominant strategy}
Energetic analysis (Fig.~\ref{energy}) suggests a systematic redistribution of positive work in the robot compared with humans: the robot produces substantially higher knee positive work while exhibiting reduced ankle positive work, indicating a knee-dominant propulsion strategy with insufficient ankle push-off. This pattern is consistent with the reduced ankle power similarity and the altered torque--angle loop structures. Such a redistribution may be an emergent solution that satisfies stability and tracking constraints without requiring high peak ankle power. Practically, these findings highlight two complementary directions for improving humanoid locomotion: (i) hardware, e.g., feet/ankle compliance and elastic elements to enable passive energy exchange; and (ii) control, e.g., explicit objectives or constraints encouraging ankle push-off timing and coordinated inter-joint work sharing.

\subsection{Asymmetry as an emergent property without explicit constraints}
Across all reported speeds, humans maintain consistently better symmetry than the robot in both kinematic SI and work symmetry $A$, with the largest gap occurring at low speeds. This suggests that even when a policy produces visually plausible gaits, subtle left--right inconsistencies can persist in both trajectory extrema and energetic distribution. A likely explanation is that RL/IL optimization can converge to any stable limit cycle that satisfies task rewards, and symmetry is not guaranteed unless it is explicitly encoded (e.g., via symmetry losses, mirrored data augmentation, or bilateral regularization). GDAF provides quantitative targets for such regularization: for example, distal-joint SI and bilateral work symmetry can be directly minimized during training or used as selection criteria for policy deployment.


\subsection{Implications for evidence-based humanoid locomotion optimisation}
Beyond reporting divergence, the main utility of GDAF is diagnostic: it pinpoints where (which joints), what (waveform, symmetry, work), and when (which speeds, which gait phases) the robot diverges from human biomechanics. This supports an evidence-based optimisation loop: use GDAF to identify dominant failure modes (e.g., low-speed work asymmetry, distal-joint kinetic mismatch), then refine reward terms, add symmetry constraints, adjust foot/ankle design, or incorporate compliance, and re-evaluate using the same speed-continuous benchmark. In this sense, GDAF complements qualitative video inspection by providing interpretable, reproducible metrics that can guide both controller development and hardware iteration.

\section{Conclusion}
This study presented GDAF, a unified method for quantitatively comparing human and humanoid gait biomechanics. By introducing a standardized processing pipeline and a comprehensive divergence metric, GDAF enables systematic evaluation of joint kinematics and kinetics across multiple walking speeds. We further established a reproducible speed-continuous benchmark and provided accompanying analysis tools. The proposed framework offers a quantitative basis for evaluating and benchmarking humanoid locomotion. Future work will extend GDAF to additional locomotion tasks and integrate its metrics into controller optimization to support the development of more biomechanically consistent humanoid gait.

\section*{Acknowledgment}
The authors would like to express their gratitude to Zhewei Shen and Xuetao Chen for their assistance in data collection and processing.



\bibliographystyle{IEEEtran}
\bibliography{reference.bib}

@article{camargo2021comprehensive,
  title={A comprehensive, open-source dataset of lower limb biomechanics in multiple conditions of stairs, ramps, and level-ground ambulation and transitions},
  author={Camargo, Jonathan and Ramanathan, Aditya and Flanagan, Will and Young, Aaron},
  journal={Journal of Biomechanics},
  volume={119},
  pages={110320},
  year={2021},
  publisher={Elsevier}
}

@article{kagami2004measurement,
  title={Measurement and comparison of humanoid H7 walking with human being},
  author={Kagami, Satoshi and Mochimaru, Masaaki and Ehara, Yoshihiro and Miyata, Natsuki and Nishiwaki, Koichi and Kanade, Takeo and Inoue, Hirochika},
  journal={Robotics and Autonomous Systems},
  volume={48},
  number={4},
  pages={177--187},
  year={2004},
  publisher={Elsevier}
}

@article{meng2018bipedal,
  title={Bipedal robotic walking control derived from analysis of human locomotion},
  author={Meng, Lin and Macleod, Catherine A and Porr, Bernd and Gollee, Henrik},
  journal={Biological cybernetics},
  volume={112},
  pages={277--290},
  year={2018},
  publisher={Springer}
}

@article{delp2007opensim,
  title={OpenSim: open-source software to create and analyze dynamic simulations of movement},
  author={Delp, Scott L and Anderson, Frank C and Arnold, Allison S and Loan, Peter and Habib, Ayman and John, Chand T and Guendelman, Eran and Thelen, Darryl G},
  journal={IEEE transactions on biomedical engineering},
  volume={54},
  number={11},
  pages={1940--1950},
  year={2007},
  publisher={IEEE}
}

@article{lee2008biomechanics,
  title={Biomechanics of overground vs. treadmill walking in healthy individuals},
  author={Lee, Song Joo and Hidler, Joseph},
  journal={Journal of applied physiology},
  volume={104},
  number={3},
  pages={747--755},
  year={2008},
  publisher={American Physiological Society}
}

@article{seo2025fasttd3,
  title={Fasttd3: Simple, fast, and capable reinforcement learning for humanoid control},
  author={Seo, Younggyo and Sferrazza, Carmelo and Geng, Haoran and Nauman, Michal and Yin, Zhao-Heng and Abbeel, Pieter},
  journal={arXiv preprint arXiv:2505.22642},
  year={2025}
}

@inproceedings{singh2022learning,
  title={Learning bipedal walking on planned footsteps for humanoid robots},
  author={Singh, Rohan P and Benallegue, Mehdi and Morisawa, Mitsuharu and Cisneros, Rafael and Kanehiro, Fumio},
  booktitle={2022 IEEE-RAS 21st International Conference on Humanoid Robots (Humanoids)},
  pages={686--693},
  year={2022},
  organization={IEEE}
}

@article{lee2025phuma,
  title={Phuma: Physically-grounded humanoid locomotion dataset},
  author={Lee, Kyungmin and Kim, Sibeen and Park, Minho and Kim, Hyunseung and Hwang, Dongyoon and Lee, Hojoon and Choo, Jaegul},
  journal={arXiv preprint arXiv:2510.26236},
  year={2025}
}

@inproceedings{wehner2009optimizing,
  title={Optimizing the Gait of a Humanoid Robot Towards Human-like Walking.},
  author={Wehner, Sven and Bennewitz, Maren},
  booktitle={ECMR},
  pages={277--282},
  year={2009}
}

@article{ji2021simulation,
  title={Simulation analysis of impulsive ankle push-Off on the walking speed of a planar biped robot},
  author={Ji, Qiaoli and Qian, Zhihui and Ren, Lei and Ren, Luquan},
  journal={Frontiers in Bioengineering and Biotechnology},
  volume={8},
  pages={621560},
  year={2021},
  publisher={Frontiers Media SA}
}

@book{winter2009biomechanics,
  title={Biomechanics and motor control of human movement},
  author={Winter, David A},
  year={2009},
  publisher={John wiley \& sons}
}

@article{patterson2012gait,
  title={Gait symmetry and velocity differ in their relationship to age},
  author={Patterson, Kara K and Nadkarni, Neelesh K and Black, Sandra E and McIlroy, William E},
  journal={Gait \& posture},
  volume={35},
  number={4},
  pages={590--594},
  year={2012},
  publisher={Elsevier}
}

@article{alves2020quantifying,
  title={Quantifying asymmetry in gait: the weighted universal symmetry index to evaluate 3d ground reaction forces},
  author={Alves, S{\'o}nia A and Ehrig, Rainald M and Raffalt, Peter C and Bender, Alwina and Duda, Georg N and Agres, Alison N},
  journal={Frontiers in bioengineering and biotechnology},
  volume={8},
  pages={579511},
  year={2020},
  publisher={Frontiers Media SA}
}

@article{winter1983energy,
  title={Energy generation and absorption at the ankle and knee during fast, natural, and slow cadences},
  author={Winter, David A},
  journal={Clinical Orthopaedics and Related Research{\textregistered}},
  volume={175},
  pages={147--154},
  year={1983},
  publisher={LWW}
}

@article{rouse2014clutchable,
  title={Clutchable series-elastic actuator: Implications for prosthetic knee design},
  author={Rouse, Elliott J and Mooney, Luke M and Herr, Hugh M},
  journal={The International Journal of Robotics Research},
  volume={33},
  number={13},
  pages={1611--1625},
  year={2014},
  publisher={SAGE Publications Sage UK: London, England}
}

\newpage
\onecolumn
\begin{appendices}
\appendix
\setcounter{table}{0}
\setcounter{figure}{0}
\renewcommand{\thetable}{A.\Roman{table}}
\renewcommand{\thefigure}{A.\arabic{figure}}

\subsection{Overview of Joint Kinematics and Kinetics}

This figure summarizes the lower joint-level kinematic and kinetic profiles for the human subject and the humanoid robot across 28 walking speeds.
\begin{figure*}[!h]
    \centering
    \begin{subfigure}{0.48\linewidth}
        \includegraphics[width=\linewidth]{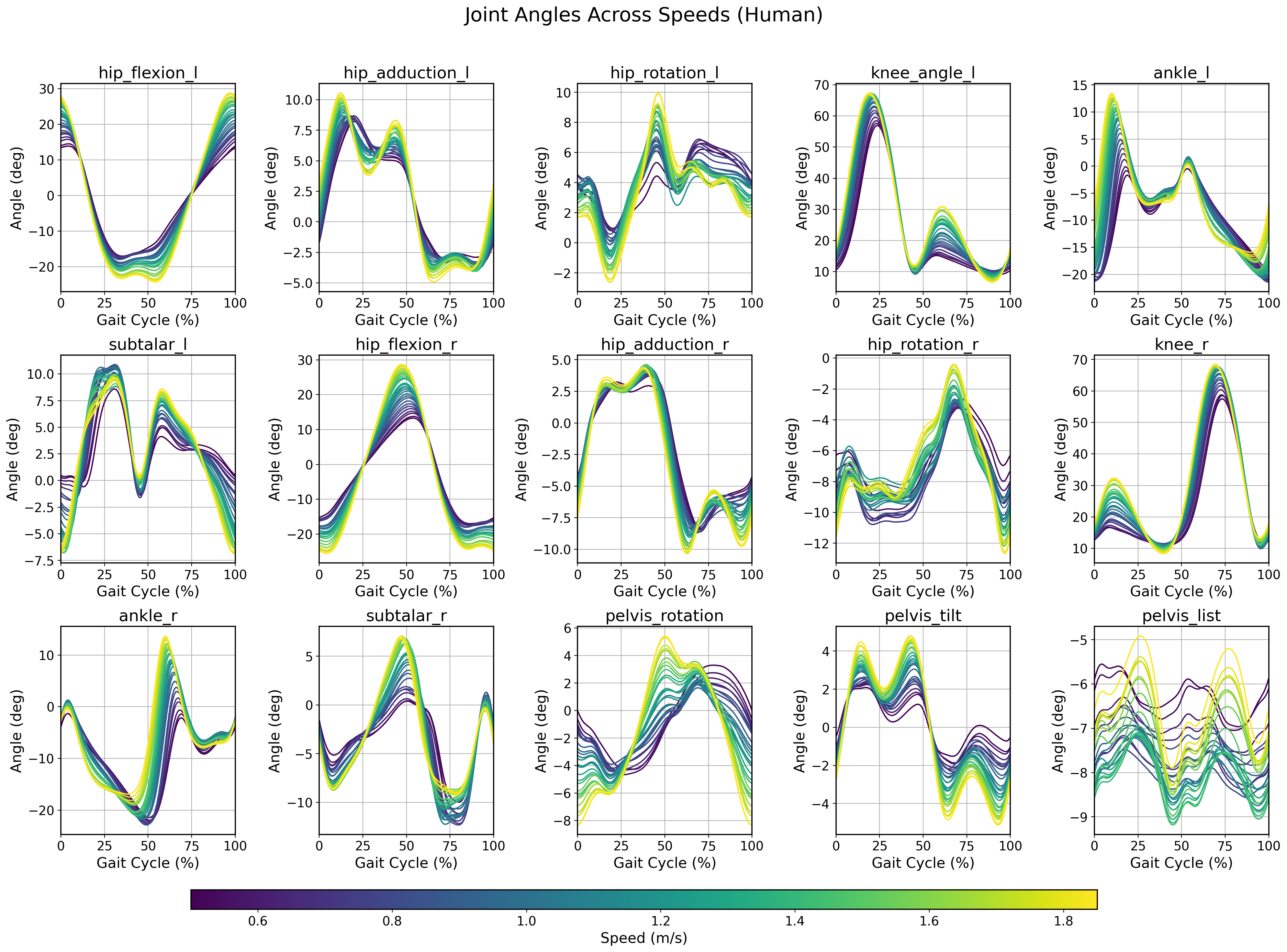}
        \caption{Joint angles from human.}
        \label{Human_Joint_Angles}
    \end{subfigure}
    \begin{subfigure}{0.48\linewidth}
        \includegraphics[width=\linewidth]{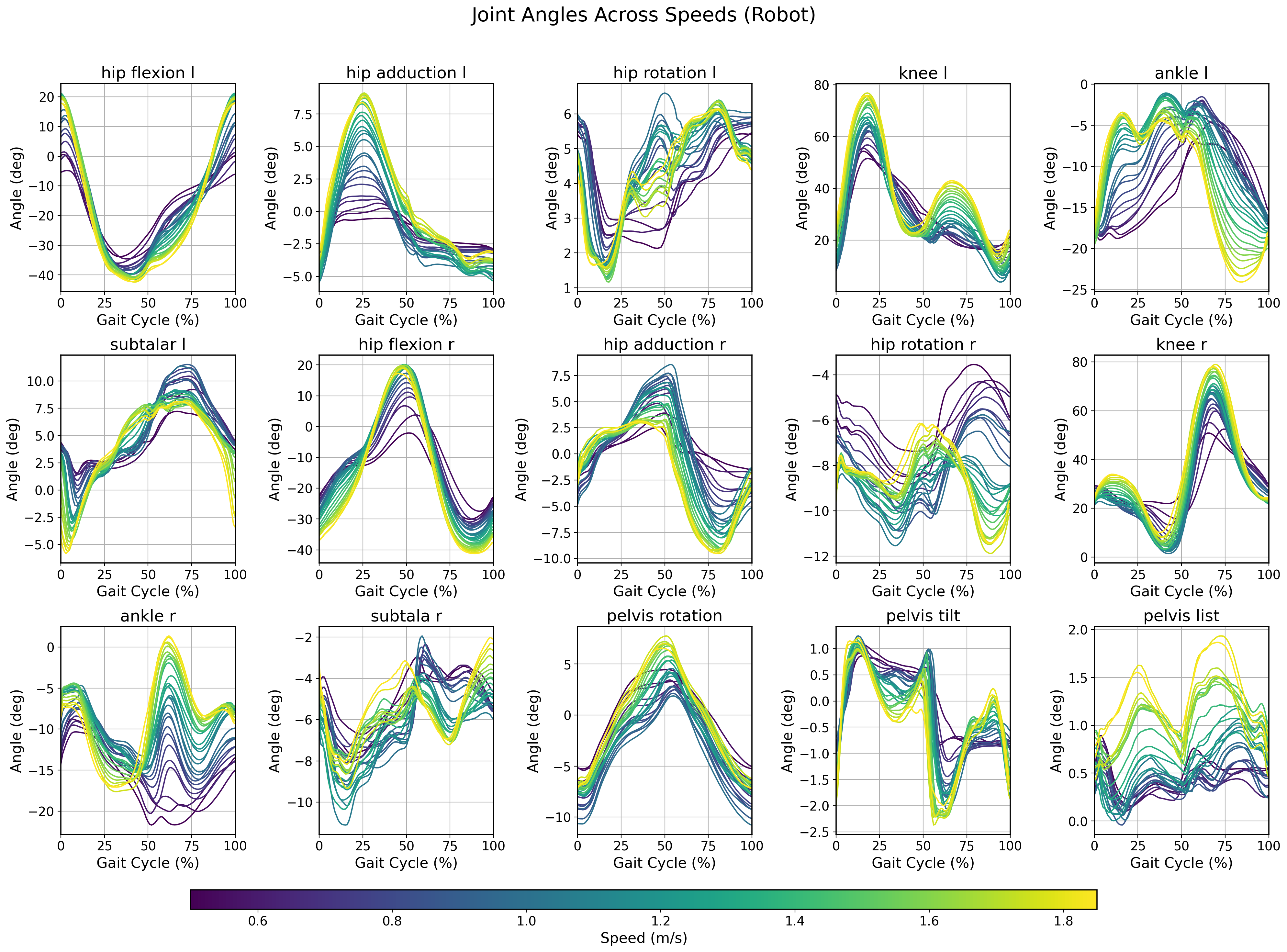}
        \caption{Joint angles from humanoid robot.}
        \label{Robot_Joint_Angles}
    \end{subfigure}

    \begin{subfigure}{0.48\linewidth}
        \includegraphics[width=\linewidth]{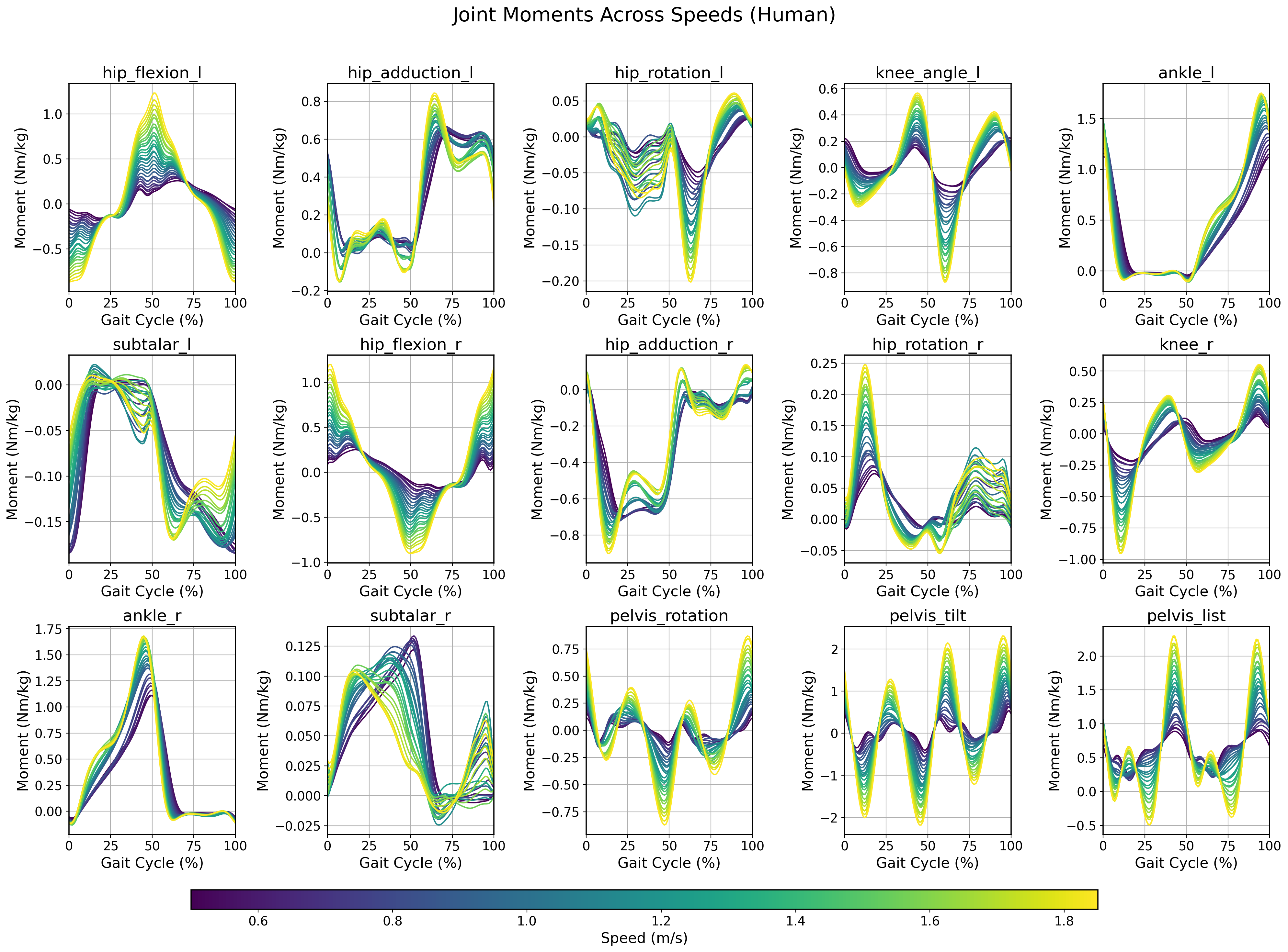}
        \caption{Joint moments from human.}
        \label{Human_Joint_Moments}
    \end{subfigure}
    \begin{subfigure}{0.48\linewidth}
        \includegraphics[width=\linewidth]{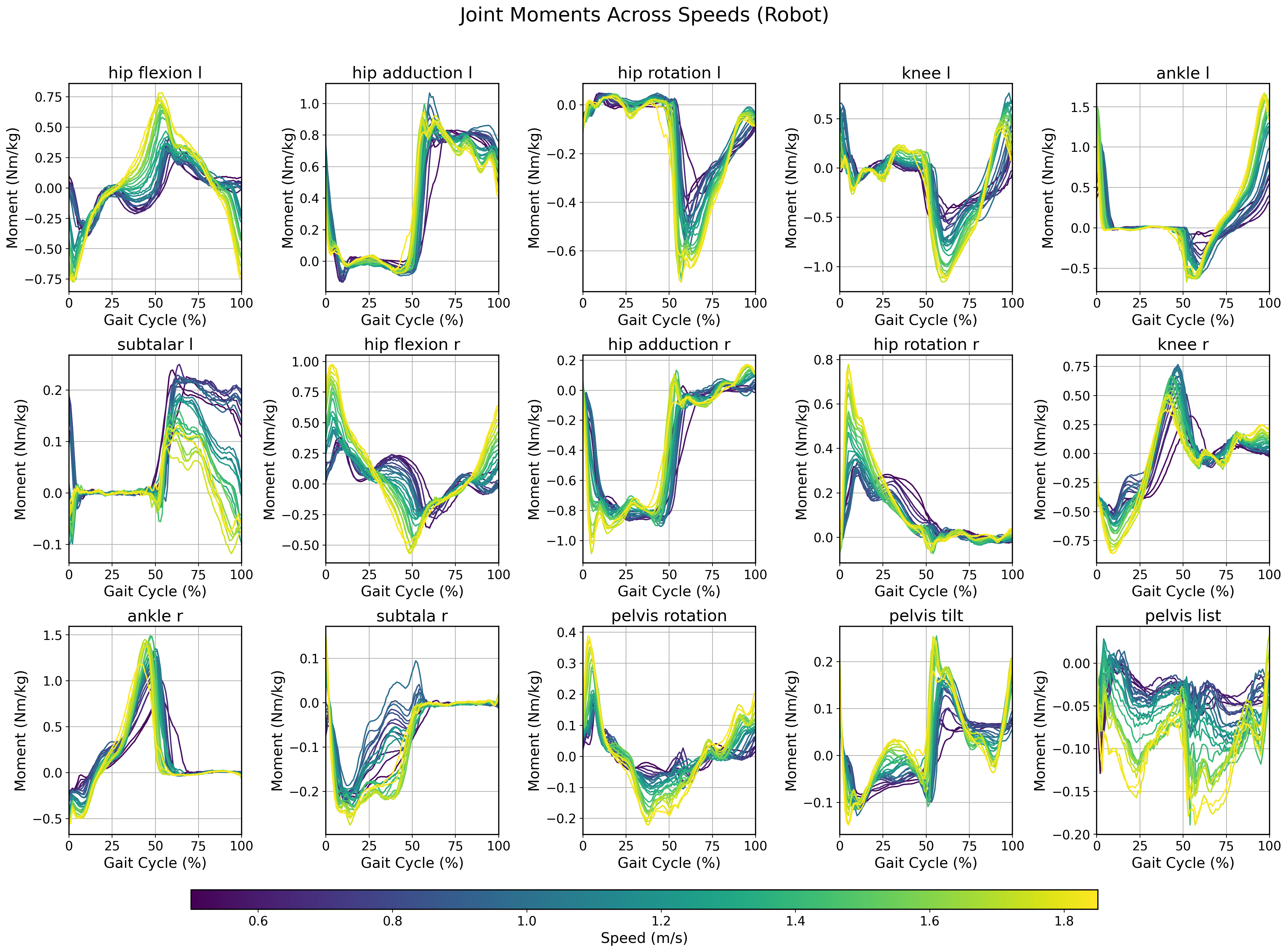}
        \caption{Joint moments from humanoid robot.}
        \label{Robot_Joint_Moments}
    \end{subfigure}

    \begin{subfigure}{0.48\linewidth}
        \includegraphics[width=\linewidth]{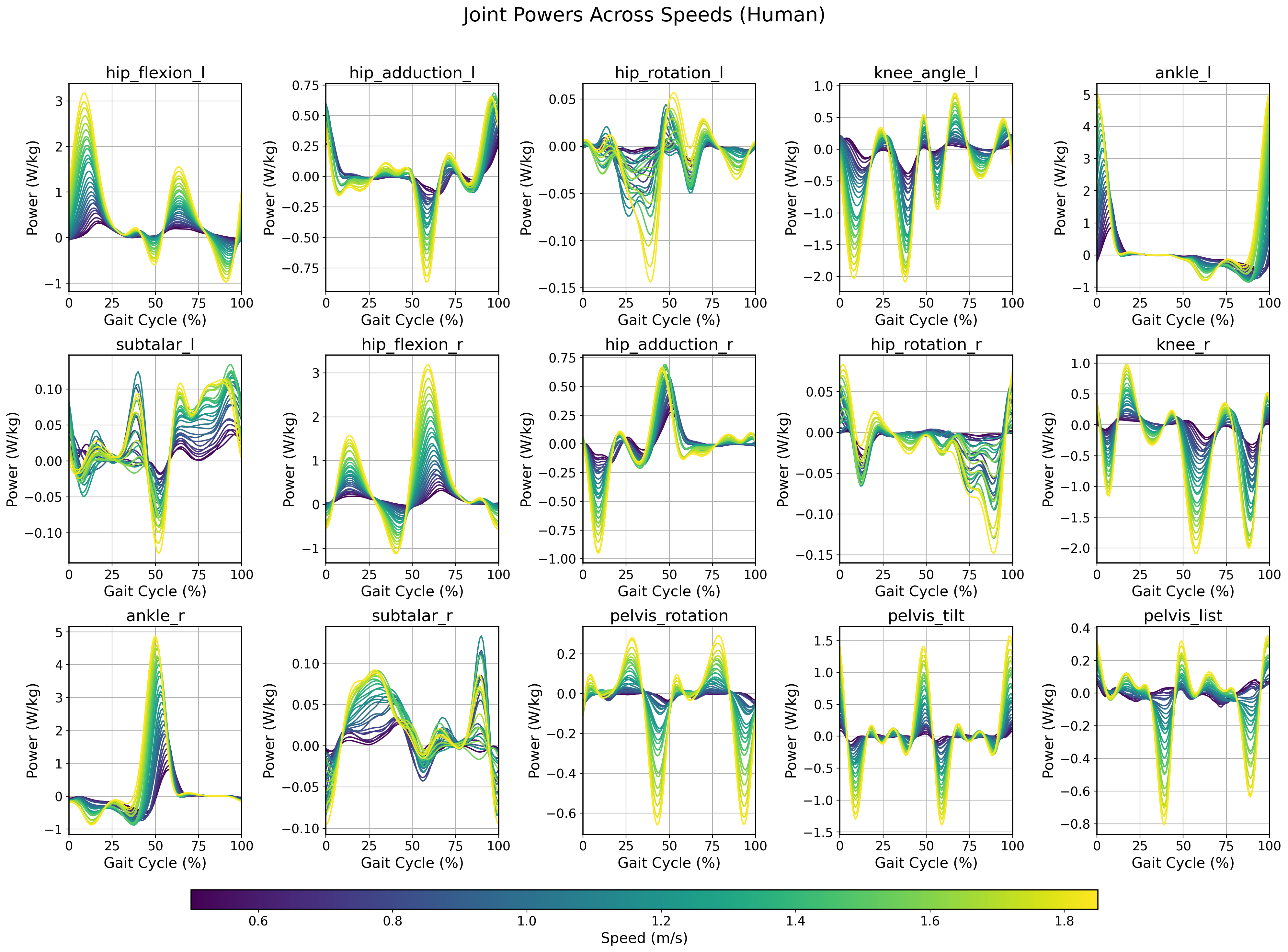}
        \caption{Joint powers from human.}
        \label{Human_Joint_Powers}
    \end{subfigure}
    \begin{subfigure}{0.48\linewidth}
        \includegraphics[width=\linewidth]{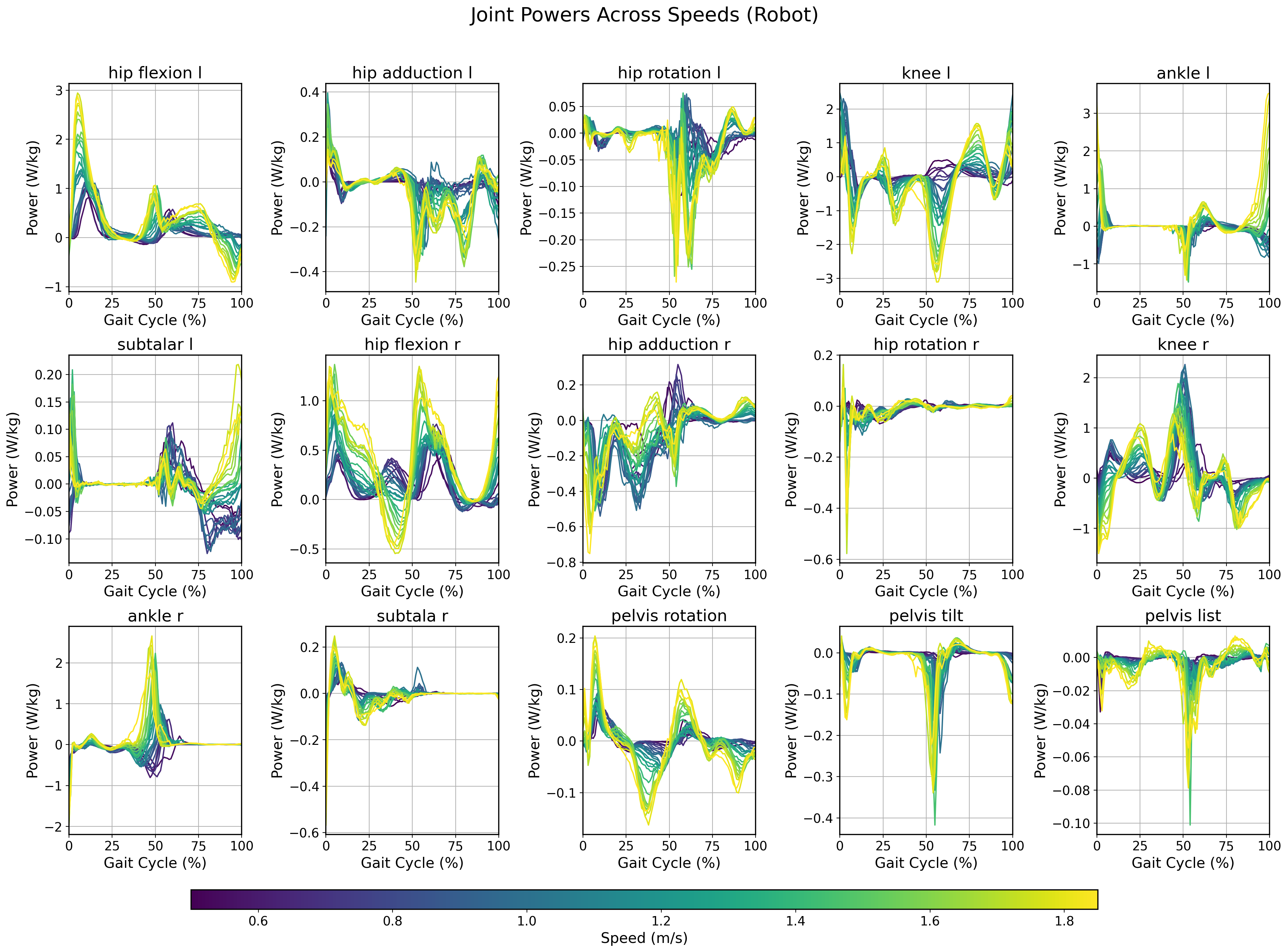}
        \caption{Joint powers from humanoid robot.}
        \label{Robot_Joint_Powers}
    \end{subfigure}

    \caption{Joint angles, moments, and power between human subject and humanoid robot under 28 speeds.}
    \label{fig:human_robot_joint_comparison}
\end{figure*}
\newpage

\subsection{MuJoCo-based Gait Visualization Tool}
\begin{figure*}[!h]
    \centering
    \includegraphics[width=\linewidth]{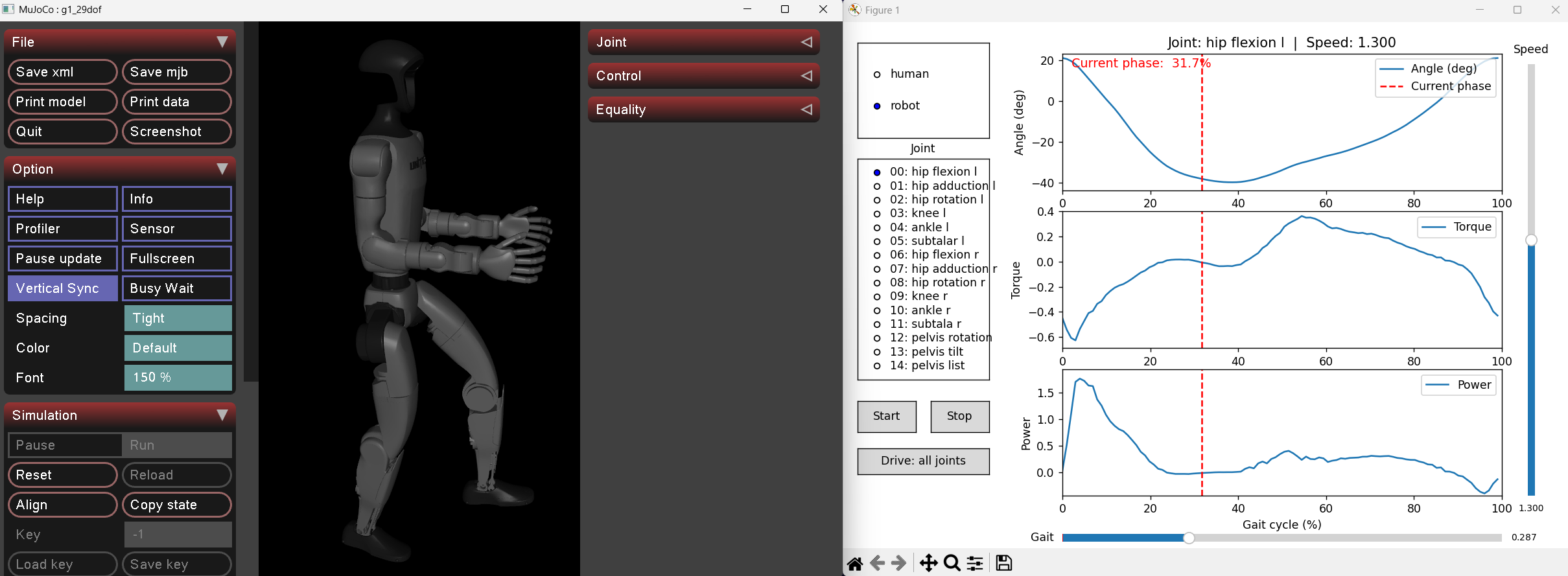}
    \caption{MuJoCo-based gait visualization tool. A 29-DOF Unitree G1 model is driven by recorded human or robotic joint trajectories. The right panel shows joint kinematics and kinetics over the gait cycle, synchronized with the simulated motion.}
    \label{fig:mujoco_gait_viewer}
\end{figure*}
To better understand and debug the mapping between human and robotic gait patterns, we developed an interactive MuJoCo-based visualization tool implemented in Python (\texttt{mujoco\_gait\_player.py}). 
The viewer takes as input the preprocessed gait datasets
\texttt{processed\_data\_human.mat} and \texttt{processed\_data\_robot.mat}, which contain joint position, torque, and power trajectories over the gait cycle for multiple walking speeds.

These trajectories are mapped to a 29-DOF Unitree G1 model described by 
\texttt{unitree\_robots/g1/g1\_29dof.xml}.
The floating base (pelvis) is fixed so that only articulated joints move during playback.
The tool provides a split interface:
(i) a MuJoCo window rendering the 3D G1 robot, and
(ii) a Matplotlib-based control and plotting panel (Fig.~\ref{fig:mujoco_gait_viewer}).

\subsubsection{Control Interface}

The control panel exposes the following elements:
\begin{itemize}
  \item \textbf{Source selector (human / robot):} toggles between replaying human- or robot-derived trajectories from the corresponding \texttt{.mat} file.
  \item \textbf{Speed slider (vertical):} selects among all available walking speeds. 
        Each discrete slider position is annotated with the corresponding physical speed value.
        Changing the speed (when playback is stopped) immediately updates both the plotted trajectories and the robot pose at the current gait phase.
  \item \textbf{Joint selector:} a list of joints (indexed and named) that controls which joint’s curves are shown in the plots.
  \item \textbf{Play controls:} \texttt{Start} and \texttt{Stop} buttons to start or stop looping playback of the gait cycle.
  \item \textbf{Drive mode toggle:} a button that switches between driving all mapped joints (Drive: all joints) and driving only the currently selected joint (Drive: selected only), which is useful for isolating individual joint behavior.
  \item \textbf{Gait phase slider (horizontal):} when playback is stopped, this slider allows manual scrubbing through the gait cycle in the range $[0,1]$. 
        Moving the slider updates the vertical phase indicator in the plots and the robot pose in MuJoCo, enabling frame-by-frame inspection.
\end{itemize}

\subsubsection{Synchronized Plots}

The right-hand side of the control panel displays three synchronized plots for the selected joint:
\begin{itemize}
  \item joint angle vs.\  gait cycle;
  \item joint torque vs.\ gait cycle;
  \item joint power vs.\ gait cycle.
\end{itemize}

A vertical red line indicates the current gait phase, which is shared with the MuJoCo simulation.
During playback, this phase advances automatically; when playback is stopped, it is controlled by the gait phase slider.

\subsubsection{Use in This Work}

In this work, the tool is mainly used for:
\begin{itemize}
  \item visually validate the correctness of the human-to-robot joint mapping and the associated sign conventions,
  \item qualitatively compare the overall gait patterns and the changes in individual joint trajectories across different walking speeds for humans and the robot.
\end{itemize}

Although reproducing the numerical results in this paper does not rely on this visualization tool, it provides an intuitive pre-check before users process their own data, allowing them to verify joint ordering and coordinate transformations. This, in turn, helps users interpret the motion profiles of each joint and build qualitative intuition about different gait patterns.

\end{appendices}
\end{document}